\documentclass[letterpaper, 10 pt, conference]{ieeeconf}  % Comment this line out if you need a4paper

\IEEEoverridecommandlockouts                              % This command is only needed if 
\usepackage{etoolbox} % fix space between abstract header and abstract body
\usepackage[utf8]{inputenc} % allow utf-8 input
\usepackage[T1]{fontenc}    % use 8-bit T1 fonts
\usepackage[english]{babel} %for hyphenation rules
\usepackage{comment}
\usepackage{cancel}
\usepackage{csquotes}
\usepackage{listings}
\usepackage{nameref}
\usepackage{paralist}
\usepackage{xspace}
\usepackage{setspace}
\usepackage{makeidx}
\usepackage{pdfpages}
\usepackage{upgreek}
\usepackage{flushend}

\usepackage{amsmath,amssymb} % define this before the line numbering.
\usepackage{amsfonts}       % blackboard math symbols
\usepackage{mathtools}
\usepackage{array} %for table entries to be in center of cell
\usepackage{nicefrac}       % compact symbols for 1/2, etc.
\usepackage{breqn}          % dmath multiline
\usepackage{textcomp}
\usepackage{bm} %bold symbols in math mode
\usepackage{pifont}
\usepackage[
  separate-uncertainty = true,
  multi-part-units = repeat
]{siunitx}

\usepackage{graphicx}
\usepackage{adjustbox} %add additional box operations in addition to resizebox
\usepackage{epsfig}
\usepackage{float}
\usepackage{subcaption}
\usepackage[font=small,labelfont=bf]{caption}
\usepackage{lscape}      %Useful for wide tables or figures.

\usepackage{tikz}  %make graphs
\usepackage{makecell}
\usepackage{overpic}
\usepackage{algorithm}
\usepackage[noend]{algpseudocode}

\usepackage{array} %for table entries to be in center of cell
\usepackage{tabularx}
\usepackage{multirow}
\usepackage{multicol}
\usepackage{booktabs}   % Publication quality tables
\usepackage{paralist}
\let\labelindent\relax %already included in IEEE
\usepackage{enumitem}
\setlist[itemize]{noitemsep,leftmargin=*,topsep=0in}
\setlist[enumerate]{noitemsep,leftmargin=*,topsep=0in}

\usepackage{url}            % simple URL typesetting
\usepackage{color,colortbl} %superceded by xcolor
\usepackage{soul}

\PassOptionsToPackage{capitalize, noabbrev}{cleveref}
\usepackage{hyperref}
\hypersetup{pagebackref=false,breaklinks=true,colorlinks,urlcolor=blue,citecolor=blue,linkcolor=blue,bookmarks=false}

\makeatletter
\let\NAT@parse\undefined
\makeatother
\usepackage[numbers,sort&compress]{natbib}

\definecolor{selfbg}{RGB}{211,228,204}
\definecolor{pursuersbg}{RGB}{227,202,203}
\definecolor{evadersbg}{RGB}{200,200,224}
\definecolor{obstaclesbg}{RGB}{187,189,192}

\newcolumntype{C}[1]{>{\centering\arraybackslash}m{#1}}
\colorlet{mycolor}{blue}

\title{\LARGE \bf
Local-Canonicalization Equivariant Graph Neural Networks for Sample-Efficient and Generalizable Swarm Robot Control
}

\author{Keqin Wang$^{*\dagger 1}$, Tao Zhong$^{*1}$, David Chang$^{1}$, Christine Allen-Blanchette$^{\dagger 1}$% <-this % stops a space
\thanks{$^*$Equal contribution. $^{1}$Princeton University.}
\thanks{$^{\dagger}$Correspondence: {\tt\small \{keqin.wang, ca15\}@princeton.edu}}
}

\begin{document}

\maketitle
\thispagestyle{empty}
\pagestyle{empty}

%%%%%%%%%%%%%%%%%%%%%%%%%%%%%%%%%%%%%%%%%%%%%%%%%%%%%%%%%%%%%%%%%%%%%%%%%%%%%%%%
\begin{abstract}

Multi-agent reinforcement learning (MARL) policies for swarm control
often learn inefficiently and generalize poorly across coordinate
frames, team sizes, and agent roles. We introduce
\textbf{L}ocal-Canonicalization \textbf{E}quivariant
\textbf{G}raph Neural Netw\textbf{o}rks (LEGO), a modular policy
architecture that combines agent-centric canonicalization with
role-aware graph encoding. Canonicalization removes dependence on the
global coordinate frame, while transforming predicted local actions
back to the world frame produces an $E(2)$-equivariant policy.
Role-wise graph encoders provide intra-role permutation equivariance
and fixed-dimensional representations for variable-size teams. LEGO can
be paired with standard MARL algorithms; we instantiate it with MAPPO.
Across cooperative \emph{MPE Spread} and competitive
\emph{Tag-occlusion} benchmarks, LEGO-MAPPO improves sample efficiency
and task performance relative to MLP-based, graph-only,
canonicalization-only, and equivariant baselines. The learned policies
transfer without fine-tuning to unseen team sizes, maintain performance
under spatial distribution shifts, and benefit from curriculum
initialization for larger teams. In Crazyflie experiments, the policy
remains operational after one pursuer is disabled. Code is available at
\url{https://github.com/CAB-Lab-Princeton/LEGO-MARL}.

%Multi-agent reinforcement learning (MARL) has emerged as a powerful paradigm for coordinating swarms of agents in complex decision-making tasks, yet significant challenges remain. In competitive settings such as pursuer-evader tasks, simultaneous adaptation can destabilize training; non-kinetic countermeasures often fail under adverse conditions; and policies trained in one configuration rarely generalize to environments with a different number of agents. To address these issues, we propose the \textbf{L}ocal-Canonicalization \textbf{E}quivariant \textbf{G}raph Neural Netw\textbf{o}rks (LEGO) framework, which integrates seamlessly with popular MARL algorithms such as MAPPO. LEGO employs graph neural networks to capture permutation equivariance and generalization to different agent numbers, canonicalization to enforce
% $E(n)$-equivariance, \change{Euclidean equivariance,} and heterogeneous representations to encode role-specific inductive biases. Experiments on cooperative and competitive swarm benchmarks show that LEGO outperforms strong baselines and improves generalization. In real-world experiments, LEGO demonstrates robustness to varying team sizes and agent failure. \change{The code is released in \url{https://github.com/CAB-Lab-Princeton/LEGO-MARL}.}

\end{abstract}

%%%%%%%%%%%%%%%%%%%%%%%%%%%%%%%%%%%%%%%%%%%%%%%%%%%%%%%%%%%%%%%%%%%%%%%%%%%%%%%%
\section{Introduction}
% \textbf{Motivation:}
% \begin{itemize}
%     \item Capacity of a collective of agents $>$ single units
%     \item Real swarms must make safe, real‑time decisions with no global state, cluttered environments, and adversaries that deceive. How? MARL
%     \item Non‑kinetic countermeasures can fail in weather/occlusion; kinetic interception remains necessary, but must be precision, low‑collateral, and scalable.
% \end{itemize}

% \textbf{Gap:}
% \begin{itemize}
%     \item MLPs suffer from the curse of dimensionality as the number of agents increases. $\rightarrow$ lack of appropriate inductive bias $\rightarrow$ permutation equivariant, GNNs
%     \item In the real world, a control problem remains fundamentally unchanged if the entire system is translated or rotated $\rightarrow$ geometric symmetries
%     \item Agents have different roles $\rightarrow$ assuming agent homogeneity limits their practical applicability
% \end{itemize}

% \begin{table}[h]
%     \centering
%     \caption{A comparison table of existing methods. ‘$\triangle$’ means that it is designed to partially have this  property.}
%     \begin{tabular}{|l|c|c|c|c|}
%          \hline
%          Model & Geometric Equiv. & Permutation Equiv. & Hetero. Agent\\
%          \hline
%          MAPPO &  &  & \\
%          \hline
%          E2GN2 & $\triangle$ & $\checkmark$ & \\
%          \hline
%          Equiv. Graphormer & $\triangle$ & $\checkmark$ & \\
%          \hline
%          LEG (ours)& $\checkmark$ & $\checkmark$ & $\checkmark$ \\
%          \hline
%     \end{tabular}
    
%     \label{tab:placeholder}
% \end{table}

Autonomous robot swarms have potential applications in environmental
monitoring, disaster response, logistics, and agriculture
~\cite{cui2022survey}. Their scalability and resilience depend on
decentralized control policies that coordinate agents through local
interactions. Multi-agent reinforcement learning (MARL)
~\cite{cui2022survey,busoniu2008comprehensive,huh2023multi} provides a
general framework for learning such policies, but learned controllers
often scale poorly and generalize weakly beyond the configurations
encountered during training.

\begin{figure}
    \centering
    \includegraphics[width=1\linewidth]{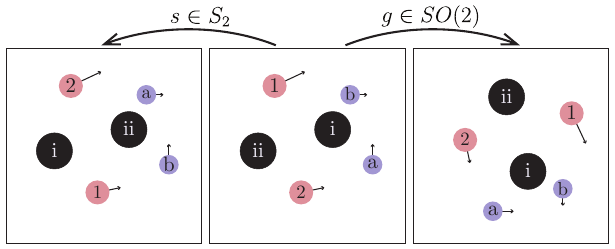}
    \caption{Permutation and rotational equivariance in the MPE
\emph{Tag} environment where \textcolor{black}{\sethlcolor{pursuersbg}\hl{pursuers}} chase \textcolor{black}{\sethlcolor{evadersbg}\hl{evaders}} while navigating \textcolor{black}{\sethlcolor{obstaclesbg}\hl{obstacles}}. . From the center configuration, swapping the
indices of two pursuers permutes their corresponding actions (left),
while rotating the scene by $90^\circ$ rotates the actions by the same
amount (right).}%An example of equivariance in the MPE Tag environment where \textcolor{black}{\sethlcolor{pursuersbg}\hl{pursuers}} chase \textcolor{black}{\sethlcolor{evadersbg}\hl{evaders}} while passing through \textcolor{black}{\sethlcolor{obstaclesbg}\hl{obstacles}}. \textit{(Middle to left)} As the agents (circles) are permuted by swapping their indices ($s\in S_2$), the optimal actions (arrows) are permuted in the same way. \textit{(Middle to right)} As the agent positions are rotated $ 90^\circ$ ($ g\in SO(2)$), the optimal actions are also rotated.}
    \label{fig:equivariance}
%\vspace{-18pt}
\end{figure}

Three structural challenges contribute to this limitation. First, the
joint state--action space grows rapidly with the number of agents,
giving rise to the ``curse of many agents''
~\cite{cui2022survey,canese2021multi}. Moreover, fixed-dimensional
policies are often tied to the number and ordering of agents used during
training, hindering transfer to different team sizes
~\cite{liu2021coach,agarwal2019learning,hu2025multitask}. Second, many
swarm-control problems exhibit task-relevant symmetries. Relabeling
agents with the same role should relabel their actions, while applying a
rigid transformation to the entire scene should transform the actions
accordingly (Figure~\ref{fig:equivariance}). Architectures that do not
encode these symmetries must instead learn them from data, reducing
sample efficiency and generalization
~\cite{van2020mdp,wang2022mathrm,nguyen2023equivariant}. Third, practical
multi-agent systems are often heterogeneous: agents may have distinct
roles or capabilities, such as pursuers and evaders
~\cite{wang2020roma,zhong2024heterogeneous}. Architectures that treat
all agents identically may therefore discard task-relevant structure.

% Furthermore, MARL algorithms are often sample inefficient. A key reason for this, is a failure to exploit task relevant symmetries inherent~\cite{van2020mdp,wang2022mathrm}. In the swarm robotics context, the optimal control strategy for tasks such as navigating to a target or surrounding an adversary, is independent of the system's absolute position and orientation (see Fig.~\ref{fig:equivariance}). However, most learning architectures do not encode this symmetry~\cite{stone2022survey}. Instead, they are trained to learn a desired behavior from scratch for every possible orientation and translation of the environment, which requires large amounts of data and training time~\cite{nguyen2023equivariant}.

% Finally, many practical applications involve agent heterogeneity~\cite{wang2020roma,zhong2024heterogeneous}; agents have distinct roles or capabilities, such as pursuers and evaders in a security task. Frameworks that assume agent homogeneity limit the complexity of learnable strategies and applicability to real-world scenarios.

To address these challenges, we introduce
\textbf{L}ocal-Canonicalization \textbf{E}quivariant
\textbf{G}raph Neural Netw\textbf{o}rks (LEGO), a modular MARL
architecture combining agent-centric canonicalization with role-aware
graph encoding. Canonicalization expresses each observation in an
$E(2)$-invariant local frame, while the graph encoder processes
variable-cardinality neighborhoods and preserves permutation symmetry
within each role. The actor predicts an action in the local frame, which
is transformed back to the global frame to obtain an $E(2)$-equivariant
policy. LEGO can be combined with standard graph encoders and MARL
optimizers; in this work, we instantiate it using Graphormer and
MAPPO~\cite{ying2021transformers,yu2022surprising}.
% In this work, we introduce the \textbf{L}ocal-Canonicalization \textbf{E}quivariant \textbf{G}raph Neural Netw\textbf{o}rks (LEGO) framework, which encodes task-relevant symmetries and agent heterogeneity directly in the policy architecture. LEGO combines agent-centric canonicalization for Euclidean equivariance with a role-aware graph encoder that provides intra-role permutation equivariance, supports variable team sizes, and preserves role-specific structure. Its modular design allows LEGO to be paired with existing MARL algorithms and instantiated with different GNN backbones~\cite{wu2020comprehensive,muller2023attending}.

The main contributions of this work are:
\begin{itemize}
    \item We introduce LEGO, a modular framework that combines agent-centric canonicalization with role-aware graph encoders to produce Euclidean- and intra-role permutation-equivariant policies for heterogeneous, variable-size teams.
    \item An empirical demonstration that LEGO, when integrated with MAPPO~\cite{yu2022surprising}, achieves superior sample efficiency and performance than strong baselines in both cooperative and competitive MARL tasks.
    \item We demonstrate zero-shot transfer to unseen team sizes, out-of-distribution generalization to unseen spatial configurations, and curriculum-based scaling to teams of up to eight agents. 
    \item We validate LEGO in a sim-to-real pursuit--evasion experiment with Crazyflie drones and show that the policy remains functional after one pursuer is disabled.\end{itemize}

\section{Related Work}
\textbf{Scalability in MARL.}
Scaling MARL to larger and variable-size teams remains challenging.
Independent learning methods~\cite{littman1994markov,tan1993multi}
treat other agents as part of the environment, which introduces
non-stationarity as their policies evolve. Centralized Training with
Decentralized Execution (CTDE)~\cite{omidshafiei2017deep,oliehoek2016concise}
mitigates this issue by exploiting global information during training
while retaining decentralized policies at execution. Representative
CTDE methods include the value-decomposition approaches
VDN~\cite{sunehag2017value} and QMIX~\cite{rashid2020monotonic}, as well
as actor--critic algorithms such as MADDPG~\cite{lowe2017multi} and
MAPPO~\cite{yu2022surprising}. Despite their effectiveness, policies
are often tied to the team size used during training. Transfer-learning
approaches~\cite{boutsioukis2011transfer,hu2025multitask} can reuse
policies across related tasks, but commonly require adaptation to the
target setting. In contrast, LEGO uses a variable-size graph
representation and role-wise pooling, allowing a single policy to be
evaluated without fine-tuning across different agent counts.

\textbf{Graph neural networks in MARL.}
Graph neural networks (GNNs) are well suited to multi-agent systems
because their shared, permutation-equivariant operations can process
variable-cardinality neighborhoods~\cite{zhao2025qmix,goeckner2024graph,
niu2021multi}. Accordingly, GNNs have been incorporated into MARL to
model interactions and communication among agents~\cite{jiang2018graph,
niu2021multi,kotecha2025leveraging}. However, standard homogeneous GNNs
do not explicitly distinguish semantic roles and are not generally
equivariant to Euclidean transformations of spatial features. Methods
such as ROMA~\cite{wang2020roma} and HARL~\cite{zhong2024heterogeneous}
address agent heterogeneity through role- or agent-specific
representations, but do not explicitly encode geometric symmetry. LEGO
combines role-aware graph encoding with agent-centric canonicalization:
the graph encoder handles within-role permutations and variable-size
neighborhoods, while canonicalization and action decanonicalization
produce an $E(2)$-equivariant policy.

\textbf{Equivariance in reinforcement learning.}
Geometric deep learning encodes task symmetries as architectural
inductive biases~\cite{bronstein2021geometric}. In reinforcement
learning, an equivariant policy maps transformed observations to
correspondingly transformed actions, which can improve sample efficiency
and generalization when the transformation preserves the underlying
decision problem~\cite{van2020mdp,wang2022mathrm}. Early equivariant
MARL methods primarily addressed permutations of agent identities
~\cite{jianye2022boosting,liu2020pic,qin2021learning,an2024scalable},
while more recent work incorporates Euclidean symmetries through
specialized equivariant actor--critic or graph architectures
~\cite{chen2023rm,mcclellan2024boosting,mcclellan2025penguin}.

Many of these approaches use steerable message passing
~\cite{brandstetter2021geometric} or $E(n)$-equivariant GNNs
~\cite{satorras2021n}, which can introduce substantial computational
overhead, and prior evaluations have largely focused on cooperative
settings. Canonicalization provides an alternative by expressing inputs
in a consistently chosen local frame
~\cite{du2022se,ma2024canonicalization,lippmann2024beyond}. LEGO adopts
this strategy to separate geometric symmetry handling from relational
encoding and policy optimization, allowing standard graph encoders such
as Graphormer~\cite{ying2021transformers} to be combined with MARL
algorithms such as MAPPO~\cite{yu2022surprising}.

\section{Preliminaries}
\textbf{Partially Observable Markov Games.}
We model $N$ agents as a partially observable Markov
game~\cite{littman1994markov,oliehoek2016concise}

\[\mathcal{M}=\langle\mathcal{I},\mathcal{S},
\{\mathcal{A}_i\},P,\{\mathcal{O}_i\},Z,\{\mathcal{R}_i\},\gamma\rangle
,\]
where $\mathcal{I}=\{1,\ldots,N\}$, $\mathcal{S}$ is the global state
space, and $\mathcal{A}_i$ and $\mathcal{O}_i$ are the action and
observation spaces of agent $i$. The kernels
$P(s'\mid s,\mathbf{a})$ and $Z(\mathbf{o}\mid s',\mathbf{a})$
govern state transitions and observations, respectively, and
$\mathcal{R}_i(s,\mathbf{a})$ is agent $i$'s reward. A general
decentralized policy conditions on the local action--observation history
$\tau_t^i$, although LEGO uses reactive actors
$a_t^i\sim\pi_i(\cdot\mid o_t^i)$. For the joint policy
$\boldsymbol{\pi}=(\pi_1,\ldots,\pi_N)$, agent $i$ maximizes
\[
J_i(\boldsymbol{\pi})
=
\mathbb{E}_{\boldsymbol{\pi},P,Z}
\left[
\sum_{t=0}^{T-1}
\gamma^t\mathcal{R}_i(s_t,\mathbf{a}_t)
\right].
\]

\textbf{Equivariance.}
Let a group $G$ act on spaces $X$ and $Y$ through
$\rho_X(g):X\rightarrow X$ and $\rho_Y(g):Y\rightarrow Y$.
A function $f:X\rightarrow Y$ is $G$-equivariant if
$f\left(\rho_X(g)x\right)
=
\rho_Y(g)f(x)$ for all $g\in G$, and $x\in X$.
Invariance is the special case in which the output action is trivial,
so that
$f(\rho_X(g)x)=f(x)$.
In this work, the relevant groups are $E(2)$ transformations of spatial
coordinates and actions, and permutations of agents within each role.
Encoding these symmetries in the architecture can improve sample
efficiency and generalization~\cite{wang2022mathrm}.

\section{Method}
% \subsection{Problem Formulation. } In this paper, we consider a swarm system in the 2D plane, where the state of each agent $i$ consists of its position $p_i \in \mathbb{R}^2$ and velocity $v_i \in \mathbb{R}^2$. Generally, the observation available to agent $i$ consists of the positions and velocities of other agents expressed in the global frame, denoting as $O_i = \{ v_i,p_i, v_j, p_i\}_{j\neq i}$. 
\begin{figure*}[!h]
    \centering
    \includegraphics[width=0.98\linewidth]{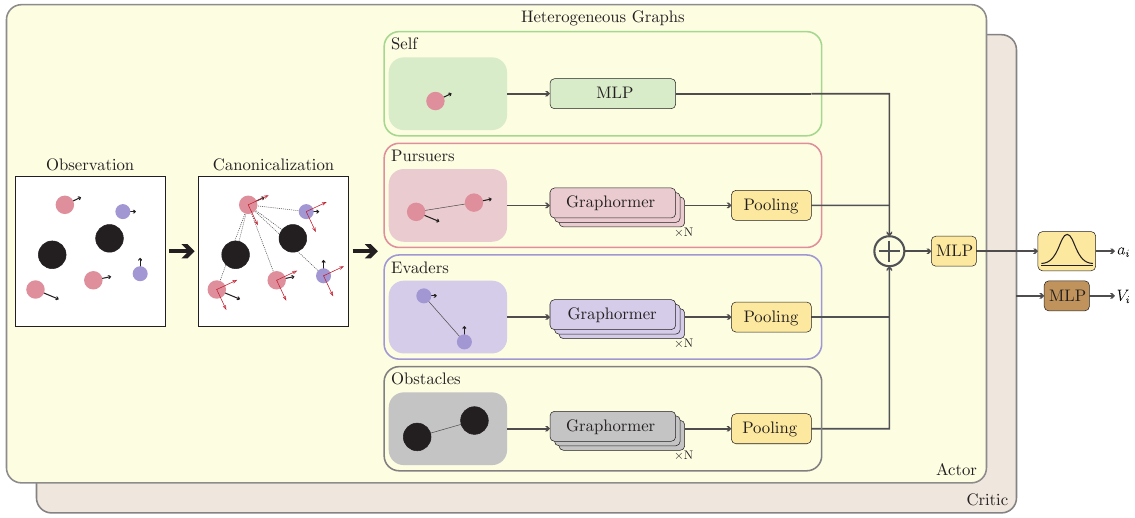}
    \caption{\textbf{LEGO architecture.}
For each agent $i$, LEGO expresses the local observation $O_i$ in a
canonical frame, partitions visible entities by role (e.g., {\sethlcolor{selfbg}\hl{self}}, \textcolor{black}{\sethlcolor{pursuersbg}\hl{pursuers}}, \textcolor{black}{\sethlcolor{evadersbg}\hl{evaders}}, \textcolor{black}{\sethlcolor{obstaclesbg}\hl{obstacles}}), and encodes and
pools each role into a fixed-dimensional representation $s_i$.
The actor predicts a local action
$a_i^{\mathrm{loc}}\sim
\pi_{\theta_{\kappa_i}}(\cdot\mid s_i)$,
which is mapped to the global frame as
$a_i=R_i a_i^{\mathrm{loc}}$.
During centralized training, an analogous encoder processes the global
state to produce the individual value estimate
$V_i=V_{\phi_{\kappa_i}}(s_i^{\mathrm{full}})$.
%\textbf{Using LEGO in MARL.} For each agent $i$, its observation $O_i$ is canonicalized as $\mathcal{C}(O_i)$ in the agent-centric frame. Role-based subgraphs (e.g., {\sethlcolor{selfbg}\hl{self}}, \textcolor{black}{\sethlcolor{pursuersbg}\hl{pursuers}}, \textcolor{black}{\sethlcolor{evadersbg}\hl{evaders}}, \textcolor{black}{\sethlcolor{obstaclesbg}\hl{obstacles}}) are encoded with Graphormer, pooled by role, and concatenated into $s_i$. The policy outputs a local action $a_i^{\text{loc}}\sim\pi_\theta(s_i)$, which is transformed back via $a_i=R_i a_i^{\text{loc}}$. Under CTDE, the same agent-centric frame and encoding pipeline are applied to the full state $X$, producing the representation $s_i^{\text{full}}$ from $\mathcal{C}_i(X)$, from which the value is estimated as $V_i=V_\phi(s_i^{\text{full}})$.
}
    \label{fig:pipeline}
    % \vspace{-18pt}
\end{figure*}
As illustrated in Figure~\ref{fig:pipeline}, LEGO canonicalizes each
agent's observation, encodes visible entities with a role-aware graph
network, predicts an action in the canonical frame, and maps that action
back to the global frame.

\subsection{Problem Formulation}

% \change{Because gravity breaks $E(3)$ symmetry in many real-world scenarios,} 
We consider a partially observable Markov game with $N$ controlled
agents moving in the plane. Let $\mathcal{V}_t$ index all entities in
the scene, including agents and task-specific objects such as landmarks
and obstacles. Each entity $j\in\mathcal{V}_t$ has position
$p_j^t\in\mathbb{R}^2$, velocity $v_j^t\in\mathbb{R}^2$, and semantic
role $\kappa_j\in\mathcal{K}$, with $v_j^t=0$ for static objects. The
global state is
\(X_t
=
\left\{
(p_j^t,v_j^t,\kappa_j)
\right\}_{j\in\mathcal{V}_t}.\)

Agent $i$ observes only the entities in its visibility set
$\mathcal{N}_i^t\subseteq\mathcal{V}_t\setminus\{i\}$:\[
O_i^t
=
\left(
p_i^t,v_i^t,\kappa_i,
\left\{
(p_j^t,v_j^t,\kappa_j)
\right\}_{j\in\mathcal{N}_i^t}
\right).\]
It selects a planar control action $a_i^t\in\mathbb{R}^2$ and receives
reward $r_i^t$. We omit time indices below for readability. Because
$O_i$ contains global-frame quantities and varies in cardinality, a
generic fixed-input actor is neither $E(2)$-equivariant by construction
nor naturally applicable to changing neighborhood sizes.

\subsection{Agent-Centric Canonicalization}
\label{sec:canonicalization}

To construct an $E(2)$-equivariant policy, we first express each
agent's observation in a canonical agent-centric coordinate frame.
This removes dependence on the arbitrary origin, orientation, and
handedness of the global coordinate system before the observation is
passed to the graph encoder. For each agent $i$, we center the local
frame at $p_i$ and construct an orthonormal basis
$R_i=[\,x_i\;\;y_i\,]\in O(2)$. The canonical $x$-axis is aligned
with the agent's velocity:
\begin{equation}
x_i =
\begin{cases}
\dfrac{v_i}{\|v_i\|}, & \|v_i\| \neq 0, \\[6pt]
x_{\mathrm{global}}, & \|v_i\| = 0,
\end{cases}
\end{equation}
where $x_{\mathrm{global}}$ denotes the $x$-axis of the global frame
and is used as a deterministic fallback for stationary agents.
%; \change{this degenerate case is excluded from our equivariance claims.}

The velocity direction determines $x_i$ but leaves two possible
orientations for the orthogonal axis. We resolve this ambiguity using
the spatial distribution of the entities visible to agent $i$. Let
\begin{equation}
c_i =
\frac{1}{|\mathcal{N}_i|+1}
\sum_{j\in\mathcal{N}_i\cup\{i\}} p_j,
\qquad
d_i=c_i-p_i,
\end{equation}
where $c_i$ is the centroid of agent $i$ and its observed neighbors,
and $d_i$ points from agent $i$ toward this centroid. Let
$J=\left[\begin{smallmatrix}0&-1\\1&0\end{smallmatrix}\right]$
denote a $90^\circ$ counterclockwise rotation. We define
\begin{equation}
y_i
=
-\operatorname{sgn}\!\big(x_i^\top Jd_i\big)Jx_i,
\qquad
R_i=
\begin{bmatrix}
x_i & y_i
\end{bmatrix}
\in O(2).
\end{equation}
The sign choice selects between the two directions orthogonal to
$x_i$ such that $y_i^\top d_i>0$. Moreover, because the sign reverses
under a reflection, $R_i$ may have determinant $+1$ or $-1$, allowing
the handedness of the local frame to transform consistently with the
scene.

The
canonicalized observation is
\begin{equation}
\label{eq:canonicalization}
\mathcal{C}(O_i)
=
\left(
v_i',
\left\{
\left(p_{j|i},v_{j|i}\right)
\right\}_{j\in\mathcal{N}_i}
\right),
\end{equation}
where
\begin{equation}
v'_i = [\|v_i\|\;\; 0]^T,\quad p_{j|i} = R_i^\top (p_j - p_i), \quad v_{j|i} = R_i^\top v_j.
\end{equation}
The semantic labels $\kappa_i$ and $\kappa_j$ are unchanged by canonicalization and omitted from Equation~\ref{eq:canonicalization} for brevity. Thus, $p_{j|i}$ and $v_{j|i}$ are the position and velocity of entity
$j$ expressed in agent $i$'s canonical frame. 
% Equivalently, the
% spatial transformation can be written as
% $\mathcal{C}(O_i)=\rho_{E(2)}(g_i^{-1})O_i$; non-geometric
% attributes, such as role labels, are left unchanged.

For a rigid transformation $g=(Q,b)\in E(2)$, positions transform as
$p\mapsto Qp+b$ and vector quantities as $v\mapsto Qv$. Whenever the
canonical frame is uniquely defined,
$R_i(g\!\cdot\!O_i)=Q R_i(O_i)$, and consequently
\begin{equation}
\mathcal{C}(g\!\cdot\!O_i)=\mathcal{C}(O_i).
\end{equation}

% \change{For a rigid transformation $g=(Q,b)\in E(2)$, positions transform
% as $p\mapsto Qp+b$ and vector quantities as $v\mapsto Qv$.
% Writing transformed quantities with tildes, under the above conditions
% we have $\widetilde R_i=QR_i$. Since $Q^\top Q=I$,
% \[
% \begin{aligned}
% \widetilde p_{j|i}
% &=\widetilde R_i^\top(\widetilde p_j-\widetilde p_i)
%  =R_i^\top Q^\top Q(p_j-p_i)=p_{j|i},\\
% \widetilde v_{j|i}
% &=\widetilde R_i^\top\widetilde v_j
%  =R_i^\top Q^\top Qv_j=v_{j|i}.
% \end{aligned}
% \]
% Moreover, $\|\widetilde v_i\|=\|v_i\|$. Therefore,
% \begin{equation}
%     \mathcal{C}(g\!\cdot\!O_i)=\mathcal{C}(O_i).
% \end{equation}}

The canonicalized representation is therefore invariant to global
translations, rotations, and reflections.

\subsection{Role-Aware Graph Encoding}

For each agent $i$, we construct an ego-centric graph $\mathcal{G}_i$
from $\mathcal{C}(O_i)$. Visible entities are partitioned by semantic
role:
\[
\mathcal{V}_i^{(\mathrm{r})}
=
\left\{
j\in\mathcal{N}_i:\kappa_j=\mathrm{r}
\right\},
\qquad
\mathrm{r}\in\mathcal{K}.
\]
Each set $\mathcal{V}_i^{(\mathrm{r})}$ induces a fully connected
role subgraph $\mathcal{G}_i^{(\mathrm{r})}$. The self branch is
encoded separately from the multi-entity role subgraphs. Its input is
$v_i'$, while each visible entity $j$ has initial feature
\[
x_{j|i}^0
=
\left[
p_{j|i}^\top,\,
v_{j|i}^\top
\right]^\top .
\]

For each non-self role subgraph, we apply an $L$-layer
Graphormer-style self-attention encoder~\cite{ying2021transformers}.
For head $h$ at layer $\ell$, the attention score between nodes $u$
and $v$ is
\begin{equation}
\alpha_{uv}^{(\ell,h)}
=
\frac{
(W_Q^{(\ell,h)}x_u^\ell)^\top
(W_K^{(\ell,h)}x_v^\ell)
}{
\sqrt{d_h}
}.
\end{equation}
The node update is
\begin{equation}
\label{eq:attention_layer}
x_u^{\ell+1}
=
x_u^\ell
+
\sigma^\ell\!\left(
\bigoplus_{h=1}^{H}
\sum_{v\in\mathcal{V}_i^{(\mathrm{r})}}
\operatorname{softmax}_{v}
\!\left(\alpha_{uv}^{(\ell,h)}\right)
W_V^{(\ell,h)}x_v^\ell
\right),
\end{equation}
where $\sigma^\ell$ maps the concatenated head outputs back to the
node-feature dimension. The self branch is processed by an MLP.

After $L$ layers, permutation-invariant pooling gives
\[
s_i^{(\mathrm{r})}
=
\operatorname{pool}
\left(
\{x_u^L:u\in\mathcal{V}_i^{(\mathrm{r})}\}
\right).
\]
For an empty role subgraph, its pooled summary is set to the zero vector. The self and role summaries are concatenated in a fixed semantic order:
\begin{equation}
s_i
=
s_i^{\mathrm{self}}
\,\|\,
s_i^{(\mathrm{r}_1)}
\,\|\cdots\|\,
s_i^{(\mathrm{r}_{|\mathcal{K}|})}.
\end{equation}
Self-attention is permutation equivariant and pooling is permutation
invariant, so $s_i$ is invariant to the ordering of entities within
each role. Its dimension depends on the number of roles rather than
the number of visible entities, allowing the same encoder to process
variable-size teams. Roles are not assumed exchangeable and are
therefore concatenated in a fixed semantic order.

\begin{algorithm}[t]
\caption{LEGO-MAPPO Training}
\label{alg:lego_mappo}
\begin{algorithmic}[1]
\Require Role-wise $\pi_{\theta_\kappa}$ and $V_{\phi_\kappa}$,
Graphormers, horizon $T$.
\For{each episode}
  \State Reset env; obtain $\{O_i^0\}_{i=1}^N$ and $X_0$.
  \For{$t=0\ldots T\!-\!1$}
    \State \textcolor{gray}{\# Canonicalize and encode}
    \For{each agent $i=1,\dots,N$}
      \State Build $R_i^t$ and compute $\mathcal C(O_i^t)$.
      \State Build role subgraphs
      $\{\mathcal G_i^{(r),t}\}_{r\in\mathcal K}$.
      \For{each role $r\in\mathcal K$}
        \State Encode/pool
        $\mathcal G_i^{(r),t}\Rightarrow s_i^{(r),t}$.
      \EndFor
      \State MLP-encode self; fuse summaries into $s_i^t$.
      \State Similarly,
$\mathcal C_i(X_t)\Rightarrow s_i^{\mathrm{full},t}$.
      \State $V_i^t=
      V_{\phi_{\kappa_i}}(s_i^{\mathrm{full},t})$.
    \EndFor
    \State \textcolor{gray}{\# Act \& step}
    \For{each agent $i=1,\dots,N$}
      \State Sample $a_i^{\mathrm{loc},t}
      \sim\pi_{\theta_{\kappa_i}}(\cdot\mid s_i^t)$.
      \State $a_i^t=R_i^t a_i^{\mathrm{loc},t}$.
    \EndFor
    \State Execute $\{a_i^t\}_{i=1}^N$; store transition.
  \EndFor
  \State \textcolor{gray}{\# CTDE update}
  \State Update all $\theta_\kappa,\phi_\kappa$ from collected trajectories.
\EndFor
\State \textbf{return}
$\{\pi_{\theta_\kappa},V_{\phi_\kappa}\}_{\kappa}$.
\end{algorithmic}
\end{algorithm}

\subsection{Equivariant Policy}

For an agent $i$ with controlled role $\kappa_i$, the actor predicts a
local action according to
\begin{equation}
a_i^{\mathrm{loc}}
\sim
\pi_{\theta_{\kappa_i}}(\cdot\mid s_i),
\end{equation}
where parameters are shared among agents with the same role. The action
is mapped back to the global frame before execution:
\begin{equation}
a_i=R_i a_i^{\mathrm{loc}}.
\label{eq:global-action}
\end{equation}

Under a global transformation $g=(Q,b)\in E(2)$, canonicalization
leaves $s_i$ unchanged while the frame transforms as $R_i\mapsto QR_i$.
The local action distribution is therefore unchanged, and the global
action distribution transforms by $Q$. Thus, the policy is
$E(2)$-equivariant whenever the canonical frame is uniquely defined.
Moreover, because the same actor is applied to all agents within a role,
the joint policy is equivariant to intra-role agent permutations.

\subsection{Centralized Training with Decentralized Execution}

We train LEGO with MAPPO~\cite{yu2022surprising} under centralized
training with decentralized execution. At execution time, each actor
uses only its local observation $O_i$. During training, the canonical frame determined from $O_i$ is also
applied to the full state $X$, yielding $\mathcal C_i(X)$, which
contains all entities available during centralized training. An analogous role-aware encoder produces $s_i^{\mathrm{full}}$,
from which the role-shared critic estimates
\begin{equation}
V_i=V_{\phi_{\kappa_i}}(s_i^{\mathrm{full}}).
\end{equation}
Because the critic is applied separately to each agent using parameters shared within each role, permuting agents within a role correspondingly permutes the
vector of value estimates. This retains agent-specific values, in
contrast to permutation-invariant centralized pooling as used by
PIC~\cite{liu2020pic}. The actor and critic are optimized using the
standard MAPPO objectives; LEGO modifies the policy and value
architectures rather than the underlying MARL optimizer.

% \change{
% \subsection{Centralized Critics}
% \label{sec:critic}

% Under CTDE, the critic has access to the full state $X$ during
% training. For each agent $i$, we apply the canonical frame
% $g_i=(R_i,p_i)$ constructed solely from its local observation $O_i$
% to the full state:
% \begin{equation}
% \label{eq:critic-canonicalization}
% \mathcal C_i(X)
% =
% \left(
% v_i',
% \left\{
% (p_{j|i},v_{j|i})
% \right\}_{j\in\mathcal V\setminus\{i\}}
% \right).
% \end{equation}
% Thus, $\mathcal C(O_i)$ and $\mathcal C_i(X)$ use the same
% agent-centric frame but differ in information scope: the former
% contains only locally observed entities, whereas the latter contains
% all entities available during centralized training.

% Applying the same role-aware graph construction, Graphormer encoder,
% and pooling operation to $\mathcal C_i(X)$ produces the full-state
% representation $s_i^{\mathrm{full}}$. The critic for role $\kappa_i$,
% whose parameters are shared among agents of that role, estimates
% \begin{equation}
% V_i=V_{\phi_{\kappa_i}}\!\left(s_i^{\mathrm{full}}\right).
% \end{equation}

% Because the frame is constructed equivariantly,
% $\mathcal C_i(X)$ is invariant to global $E(2)$ transformations whenever $O_i$ is nondegenerate.
% Moreover, applying each role-shared critic separately to its agents makes
% the collection of values $\{V_i\}_{i=1}^N$ equivariant to within-role
% agent permutations, rather than collapsing them into a single
% permutation-invariant summary as in PIC~\cite{liu2020pic}.
% }

\begin{figure*}[t]
    \centering
    \includegraphics[width=0.99\linewidth]{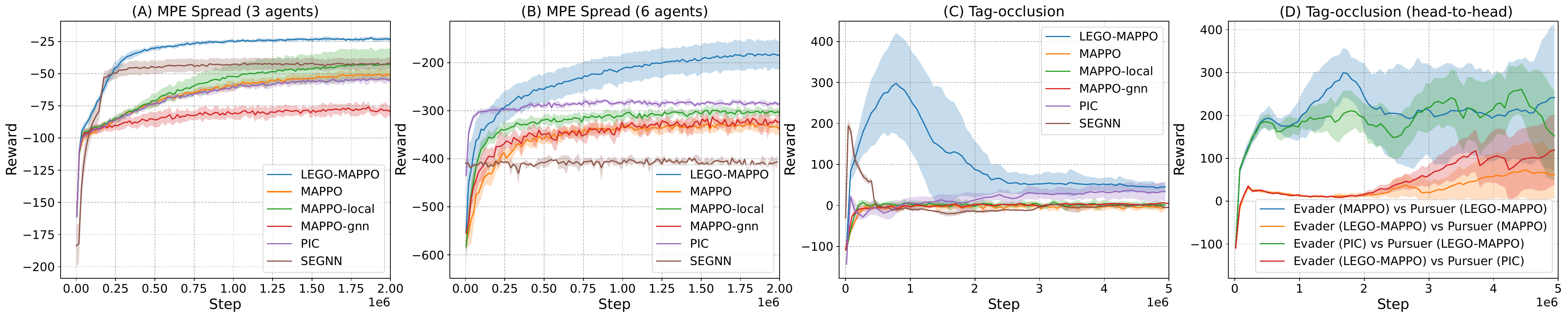}
    \caption{Comparing learning performance on MPE Spread and Tag-occlusion tasks. (A) Average episode rewards in the MPE Spread task with 3 agents and 3 landmarks. (B) Average episode rewards in the MPE Spread task with 6 agents and 6 landmarks. (C) Average episode rewards in the Tag-occlusion task with 2 evaders, 3 pursuers, and 2 obstacles. (D) Average episode reward in asymmetric head-to-head matchups.}
    \label{fig:spread}
    % \vspace{-18pt}
\end{figure*}

\section{Experiments}
We evaluate LEGO-MAPPO on cooperative and competitive benchmarks,
zero-shot transfer across team sizes, symmetry-related distribution
shifts, curriculum-based scaling, and a real-world Crazyflie experiment
in which one pursuer is disabled. Together, these experiments assess
sample efficiency, task performance, scalability, generalization, and
robustness to agent failure. Videos of the simulation and hardware
experiments are available at \url{https://lego-marl.github.io/}.

\subsection{Experimental Setup and Baselines}

We instantiate LEGO with Multi-Agent Proximal Policy Optimization
(MAPPO)~\cite{yu2022surprising}, yielding \textbf{LEGO-MAPPO}. LEGO
modifies the actor and critic architectures while retaining the standard
MAPPO optimization procedure. We compare against vanilla MAPPO, two
component ablations, and two existing MARL baselines:
\begin{itemize}
    \item \textbf{MAPPO}~\cite{yu2022surprising}: MLP-based actors and
    critics trained directly on the raw observations.

    \item \textbf{MAPPO-local}: a canonicalization-only ablation that
    expresses each observation in the agent-centric frame described in
    Section~\ref{sec:canonicalization}, while retaining MLP-based actors
    and critics.

    \item \textbf{MAPPO-GNN}: a graph-only ablation that uses LEGO's
    role-aware Graphormer encoders and pooling operations on raw,
    noncanonicalized observations.

    \item \textbf{PIC}~\cite{liu2020pic}: an MLP actor with a
    permutation-invariant GCN critic. We adapt the original
    MADDPG-based implementation to MAPPO.

    \item \textbf{SEGNN}~\cite{chen2023rm}: an actor--critic architecture
    based on steerable message passing~\cite{brandstetter2021geometric}
    that is equivariant to agent permutations and $E(2)$ transformations.
\end{itemize}
Unless otherwise noted, all methods use the same observation and action
spaces, training budget, and evaluation protocol.

\subsection{Training Performance}

\textbf{Cooperative task.}
We evaluate on the continuous-action \emph{MPE Spread}
environment~\cite{mordatch2017emergence,lowe2017multi}, in which
$N$ agents must cover $N$ landmarks while avoiding collisions. The
episode reward is the negative sum, over landmarks, of the distance to
the nearest agent, with an additional penalty of $-1$ for each
collision. The raw observations contain the agent's absolute state and
the relative states of other entities. LEGO instead expresses these
quantities in the time-varying canonical frame defined by the agent's
velocity and the centroid of its visible neighbors.

Figure~\ref{fig:spread}A--B reports average episode reward over 10
seeds for teams of three and six agents. In the three-agent setting,
LEGO-MAPPO and SEGNN converge most rapidly, while LEGO-MAPPO achieves
the highest final reward. MAPPO-local improves over vanilla MAPPO,
whereas MAPPO-GNN performs poorly despite its role-aware graph encoder,
indicating that graph structure alone does not provide the geometric
inductive bias supplied by canonicalization. In the six-agent setting,
LEGO-MAPPO converges faster and reaches substantially higher reward than
all baselines, with PIC providing the strongest competing result.
Together, comparisons with MAPPO-local and MAPPO-GNN show that neither
canonicalization nor role-aware graph encoding alone matches the
performance obtained by combining the two.

\textbf{Competitive task.}
We evaluate LEGO-MAPPO on \emph{Tag-occlusion}, adapted from MPE
Tag~\cite{terry2021pettingzoo}, in which three slower pursuers chase
two evaders while two obstacles occlude line-of-sight observations.
Figure~\ref{fig:spread}C reports the average episode reward over 10
seeds for each method, except SEGNN, which is averaged over three seeds
due to its substantially higher training cost. Because both teams learn
simultaneously, the LEGO-MAPPO episode-reward is non-monotonic: it increases
rapidly, peaks near $1\times10^6$ interactions, and then decreases and
stabilizes as the evaders improve.

\begin{figure}[t]
    \centering
    \includegraphics[width=0.95\linewidth]{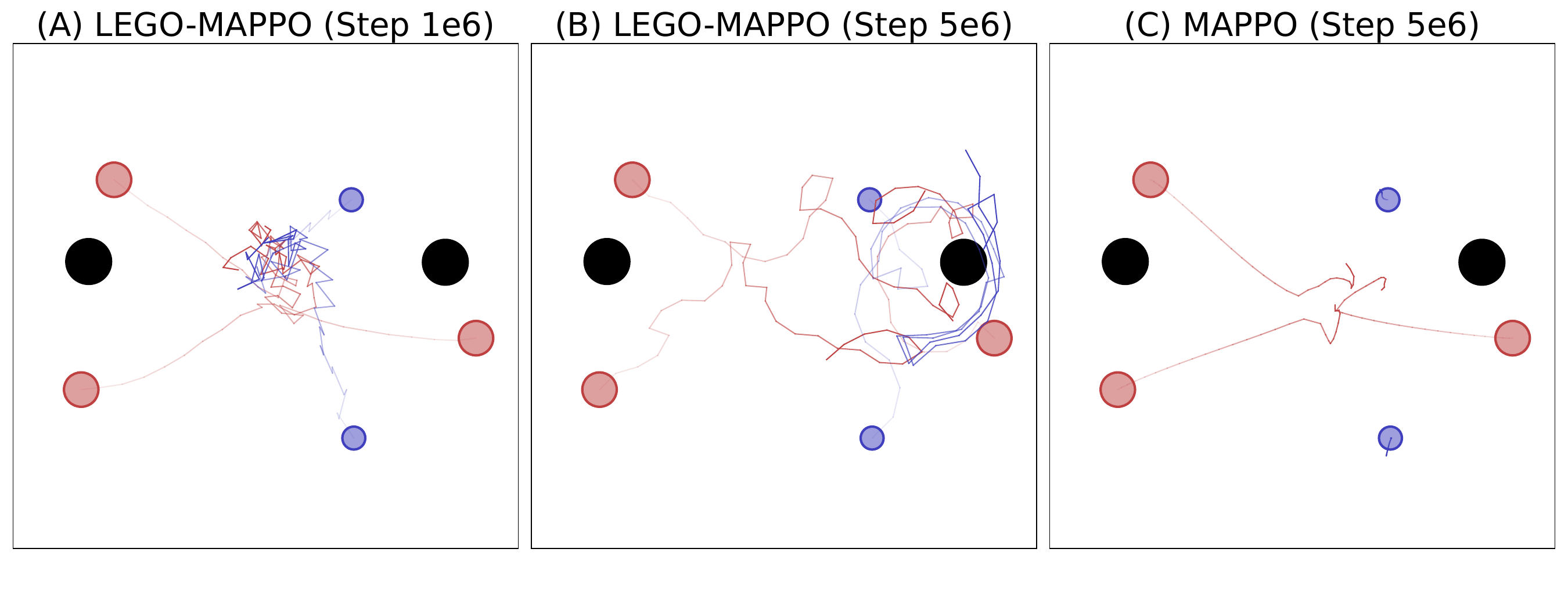}
    \caption{Example trajectories in \emph{Tag-occlusion}.
(A) LEGO-MAPPO after $1\times10^6$ training interactions;
(B) LEGO-MAPPO after $5\times10^6$ interactions; and
(C) MAPPO after $5\times10^6$ interactions.}
    \label{fig:trajectory_simulator}
    % \vspace{-18pt}
\end{figure}

The trajectories in Figure~\ref{fig:trajectory_simulator} illustrate
this co-adaptation. At $1\times10^6$ interactions, the LEGO-MAPPO
pursuers cluster near the center and form a barrier that confines the
evaders. By $5\times10^6$ interactions, the evaders exploit the
obstacles to break line of sight and sustain longer evasive trajectories,
reducing the pursuers' reward. In contrast, MAPPO does not exhibit
comparable coordinated behavior after $5\times10^6$ interactions.
SEGNN follows a similar non-monotonic trend but reaches lower rewards,
PIC improves more gradually, and the remaining baselines stay near
zero. Because self-play reward conflates improvements by both teams, we
next use head-to-head evaluation to compare the learned policies more
directly.

\textbf{Head-to-head evaluation.}
Because symmetric self-play reward reflects the simultaneous improvement
of both teams, we additionally consider asymmetric matchups in which one
team is trained with LEGO-MAPPO and the opposing team with MAPPO or PIC.
We evaluate both role assignments, using LEGO-MAPPO for either the
pursuers or the evaders. Figure~\ref{fig:spread}D reports the average
pursuer reward over 10 seeds; higher values favor the pursuers,
whereas lower values indicate more effective evasion.

\begin{figure}[t]
    \centering
    \includegraphics[width=0.95\linewidth]{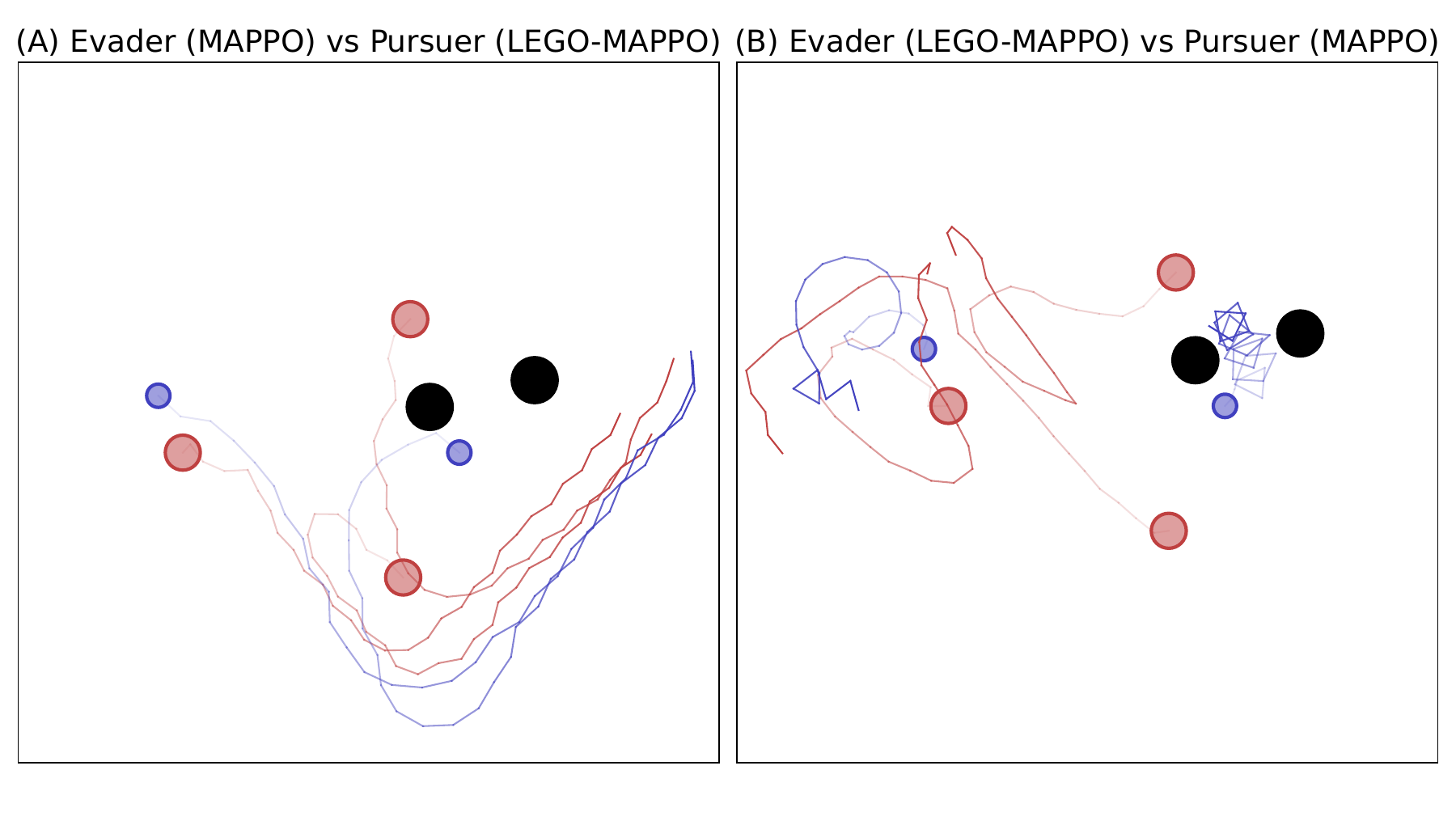}
    \caption{Example trajectories in the head-to-head evaluation. (A) LEGO-MAPPO-controlled pursuers chase MAPPO-controlled evaders. (B) MAPPO-controlled pursuers chase LEGO-MAPPO-controlled evaders.}
    \label{fig:cross_validation}
    % \vspace{-5pt}
\end{figure}

Against MAPPO, LEGO-MAPPO improves performance in both roles: LEGO-MAPPO
pursuers attain higher reward against MAPPO evaders, while MAPPO pursuers
receive lower reward against LEGO-MAPPO evaders. The trajectories in
Figure~\ref{fig:cross_validation} illustrate these behaviors. MAPPO
evaders tend to move together and are readily captured by LEGO-MAPPO
pursuers (Figure~\ref{fig:cross_validation}A), whereas LEGO-MAPPO evaders
separate and follow distinct escape routes against MAPPO pursuers
(Figure~\ref{fig:cross_validation}B). Matchups involving PIC are more
competitive, consistent with its stronger performance in the preceding
experiments.

\textbf{Run-time comparison.}
Table~\ref{tab:run_time} reports wall-clock training time for
$2\times10^6$ environment interactions on the three- and six-agent
\emph{MPE Spread} tasks, measured using an Intel Xeon Gold 5220R CPU
and an NVIDIA A6000 GPU. LEGO-MAPPO requires moderately more training
time than vanilla MAPPO, increasing from 277.9\,s to 377.6\,s for
three agents and from 814.2\,s to 997.8\,s for six agents. In contrast,
SEGNN requires 18.2 and 20.2 hours, respectively. Thus, LEGO-MAPPO
retains runtimes on the same scale as the MAPPO and PIC baselines while
remaining substantially less expensive than the steerable equivariant
baseline.

% \textbf{Run-time comparison.} Table~\ref{tab:run_time} summarizes training times for the 3-agent cooperative task over $2 \times 10^6$ steps (Intel Xeon Gold 5220R, NVIDIA A6000). SEGNN incurs exorbitant computational costs and lacks multi-threading support due to its steerable feature space. In contrast, LEGO-MAPPO achieves its significant performance gains with only a marginal computational increase, maintaining runtimes comparable to standard baselines.

\begin{table}
    \centering
    \caption{Average wall-clock training time for $2\times10^6$
environment interactions on \emph{MPE Spread}.}
    \resizebox{\columnwidth}{!}{
    \begin{tabular}{|c|c|c|c|c|}
        \hline
        Method & MAPPO & PIC & SEGNN & LEGO-MAPPO\\
        \hline
        3 agents & 277.9s & 300.5s & 18.2Hr & 377.6s \\
        \hline
        6 agents & 814.2s & 1257.3s & 20.2Hr & 997.8s\\
        \hline
    \end{tabular}}
    \label{tab:run_time}
    % \vspace{-13pt}
\end{table}

\begin{table}[t]
    \centering
    \caption{Average evaluation reward across team sizes. LEGO-MAPPO and
SEGNN are trained with four agents and evaluated without fine-tuning;
$\dagger$ denotes a policy trained specifically for the corresponding
team size.}
    \resizebox{\columnwidth}{!}{
    \begin{tabular}{|c|c|c|c|c|c|}
        \hline
        Method & LEGO-MAPPO & PIC & SEGNN &MAPPO\\
        \hline
        2 agents & -10.6$\pm$2.9  & -9.8$\pm$1.7$^\dagger$ &  -47.9$\pm$14.5&\textbf{-9.2$\pm$4.1$^\dagger$}\\
        \hline
        3 agents & \textbf{-23.8$\pm$7.4} &  -42.7$\pm$10.1$^\dagger$ &  -148.2$\pm$44.3
        & -43.4$\pm$12.1$^\dagger$ \\
        \hline
        4 agents & \textbf{-44.0$\pm$19.9$^\dagger$} &  -130.8$\pm$15.6$^\dagger$ & -84.3$\pm$39.2$^\dagger$ & -135.0$\pm$35.4$^\dagger$\\
        \hline
        5 agents& \textbf{-79.8$\pm$18.5} & -191.4$\pm$12.8$^\dagger$ & -282.9$\pm$202.7 & -216.3$\pm$43.1$^\dagger$ \\
        \hline
        6 agents& \textbf{-114.6$\pm$17.0} &  -297.3$\pm$18.5$^\dagger$ & -859.2$\pm$220.7  & -378.0$\pm$67.9$^\dagger$ \\
        \hline
    \end{tabular}}
    \label{tab:generalization}
    % \vspace{-5pt}
\end{table}

% \begin{table}[t]
%     \centering
%     \caption{Average evaluation reward result on curriculum experiment.}
%     \resizebox{\columnwidth}{!}{
%     \begin{tabular}{|c|c|c|c|c|}
%         \hline
%         Method & LEGO-MAPPO & LEGO-MAPPO-scl &LEGO-MAPPO-curr & MAPPO\\
%         \hline
%         6 agents & -196.5$\pm$39.5 & -113.4$\pm$16.3 & \textbf{-99.6$\pm$9.2} & -381.2$\pm$71.3\\
%         \hline
%         7 agents & -265.5$\pm$31.7 & -170.8$\pm$29.4 &  \textbf{-145.2$\pm$20.4} & -430.7$\pm$81.2\\
%         \hline
%         8 agents & -360.3$\pm$81.6 &  -239.7$\pm$30.1 & \textbf{-150.7$\pm$16.5}  & -508.3$\pm$54.2\\
%         \hline
%     \end{tabular}}
%     \label{tab:curriculum}
%     \vspace{-18pt}
% \end{table}

\begin{table}[t]
    \centering
    \caption{Average evaluation reward result on curriculum experiment. `–' indicates the result is unavailable due to the extremely long training time.}
    \resizebox{\columnwidth}{!}{
    \begin{tabular}{|c|c|c|c|c|}
        \hline
        Method & 6 agents & 7 agents & 8 agents \\
        \hline
        MAPPO & -381.2$\pm$71.3 & -430.7$\pm$81.2 & -508.3$\pm$54.2 \\
        \hline
        SEGNN-scl & -833.1$\pm$203.7 & -1184.4$\pm$102.7 &  -1899.6$\pm$402.8\\
        \hline
        SEGNN-curr & -186.1$\pm$40.4 & - & - \\
        \hline
        LEGO-MAPPO & -196.5$\pm$39.5 & -265.5$\pm$31.7 & -360.3$\pm$81.6 \\
        \hline
        LEGO-MAPPO-scl & -113.4$\pm$16.3 & -170.8$\pm$29.4 & -239.7$\pm$30.1 \\
        \hline
        LEGO-MAPPO-curr & \textbf{-99.6$\pm$9.2} & \textbf{-145.2$\pm$20.4} & \textbf{-150.7$\pm$16.5}  \\
        \hline
        
    \end{tabular}}
    \label{tab:curriculum_transposed}
    %\vspace{-10pt}
\end{table}

\subsection{Generalization}
\textbf{Zero-shot transfer across team sizes.}
Because LEGO pools variable-cardinality role subgraphs into a
fixed-dimensional representation, a single policy can be applied to
different numbers of agents and landmarks. We train LEGO-MAPPO on
\emph{MPE Spread} with four agents and four landmarks, and evaluate it
without fine-tuning on corresponding tasks with two, three, five, and
six agents. As shown in Table~\ref{tab:generalization}, the four-agent
policy transfers effectively to the unseen team sizes and, for three,
five, and six agents, outperforms MAPPO and PIC policies trained
specifically for each setting. Vanilla MAPPO and PIC cannot be evaluated
zero-shot because their fixed-size MLP actors depend on the training
configuration. Although SEGNN supports variable-size inputs, its
four-agent policy degrades substantially when transferred to unseen
team sizes.

% \kevin{I find a interesting thing: directly training a policy with LEGO-MAPPO in the 6-agent, 6-landmark setting yields poorer performance ($196.5\pm39.5$) than first training it in the 4-agent, 4-landmark setting and then adapting the policy to the 6-agent scenario ($-114.6\pm17.0$). I think the explanation is, because the underlying mechanism of the spread task is universal, allowing policies learned in simpler settings to transfer smoothly. In contrast, starting directly with more agents increases the search space, requiring more time to discover the underlying mechanism and often leading to sub-optimal convergence. This suggests an important takeaway for our method: a curriculum Learning strategy, where the policy is first pre-trained on simpler scenarios to capture the fundamental structure of the task and then fine-tuned in the target setting, can improve training efficiency and final performance. I plan to design a small experiment here.}

\textbf{Curriculum learning.}
We next test whether training on a smaller team provides an effective
initialization for larger-team tasks. For each target size, we compare
training from scratch with zero-shot transfer (\emph{-scl}) and
curriculum initialization (\emph{-curr}). The \emph{-scl} models are
trained with four agents and evaluated directly at the target team size,
whereas the \emph{-curr} models are pretrained for
$1\times10^6$ interactions with four agents and then fine-tuned for
$4\times10^6$ interactions with six, seven, or eight agents. We train
10 policies with independent random seeds and evaluate each policy under
10 additional environment seeds.

As shown in Table~\ref{tab:curriculum_transposed},
LEGO-MAPPO-\emph{curr} achieves the highest average reward at every
target team size. Moreover, LEGO-MAPPO-\emph{scl} outperforms training
LEGO-MAPPO from scratch, indicating that policies learned on smaller
teams provide a useful initialization for larger-team control. For
SEGNN, curriculum initialization also improves performance in the
six-agent setting, while results for larger teams are unavailable due
to its computational cost.

\begin{figure}
    % \vspace{-10pt}
    % \vspace{-1.5em}
    \centering
    \includegraphics[width=\linewidth]{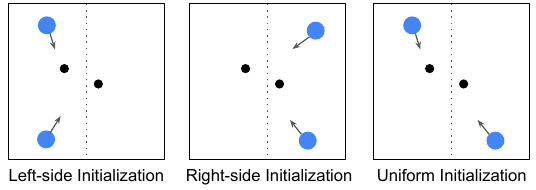}
    \caption{The illustration of the out-of-distribution task, where the agents (blue circle) need to cover the landmarks (black circle) without collision.}
    \label{fig:ood_illustration}
    % \vspace{-10pt}
\end{figure}

\textbf{Out-of-distribution generalization.}
We evaluate generalization across spatial initialization distributions
in \emph{MPE Spread}. Policies are trained with agents initialized on
the left side of the environment and evaluated on left-side, reflected
right-side, and uniform initializations, as illustrated in
Figure~\ref{fig:ood_illustration}. The right-side setting tests transfer
under a reflection of the training configuration, while the uniform
setting introduces a broader positional distribution. We compare
LEGO-MAPPO with the geometrically equivariant MAPPO-local and SEGNN
baselines, as well as vanilla MAPPO. Results are reported in
Table~\ref{tab:ood}.

LEGO-MAPPO, MAPPO-local, and SEGNN maintain similar performance across
the three initialization distributions, whereas MAPPO degrades
substantially under the right-side and uniform shifts. LEGO-MAPPO
achieves the highest reward in each setting, indicating that its
symmetry-aware architecture improves robustness to spatial distribution
shift while retaining strong in-distribution performance.

\begin{table}
    \centering
    \caption{Out-of-distribution reward result.}
    \resizebox{\columnwidth}{!}{
    \begin{tabular}{|c|c|c|c|}
        \hline
        \multirow{2}{*}{Method} & Training initialization & \multicolumn{2}{c|}{Testing initialization} \\
        \cline{2-4}
        & left-side & right-side & uniform \\
        \hline
        LEGO-MAPPO & \textbf{-21.4$\pm$6.1} & \textbf{-20.2$\pm$5.3} & \textbf{-22.03$\pm$6.7} \\
        \hline
        MAPPO-local & -39.3$\pm$17.6& -40.3$\pm$11.4 & -42.1$\pm$21.4\\
        \hline
        SEGNN & -45.3$\pm$12.9 &  -49.2$\pm$15.7 & -44.7$\pm$10.9\\
        \hline
        MAPPO & -50.7$\pm$12.1 & -72.7$\pm$20.0 & -70.9$\pm$18.4\\
        \hline
        
    \end{tabular}
    }
    \label{tab:ood}
    %\vspace{-10pt}
\end{table}

\subsection{Real-World Demonstration}

\begin{figure}[t]
    \centering
    \includegraphics[width=\linewidth]{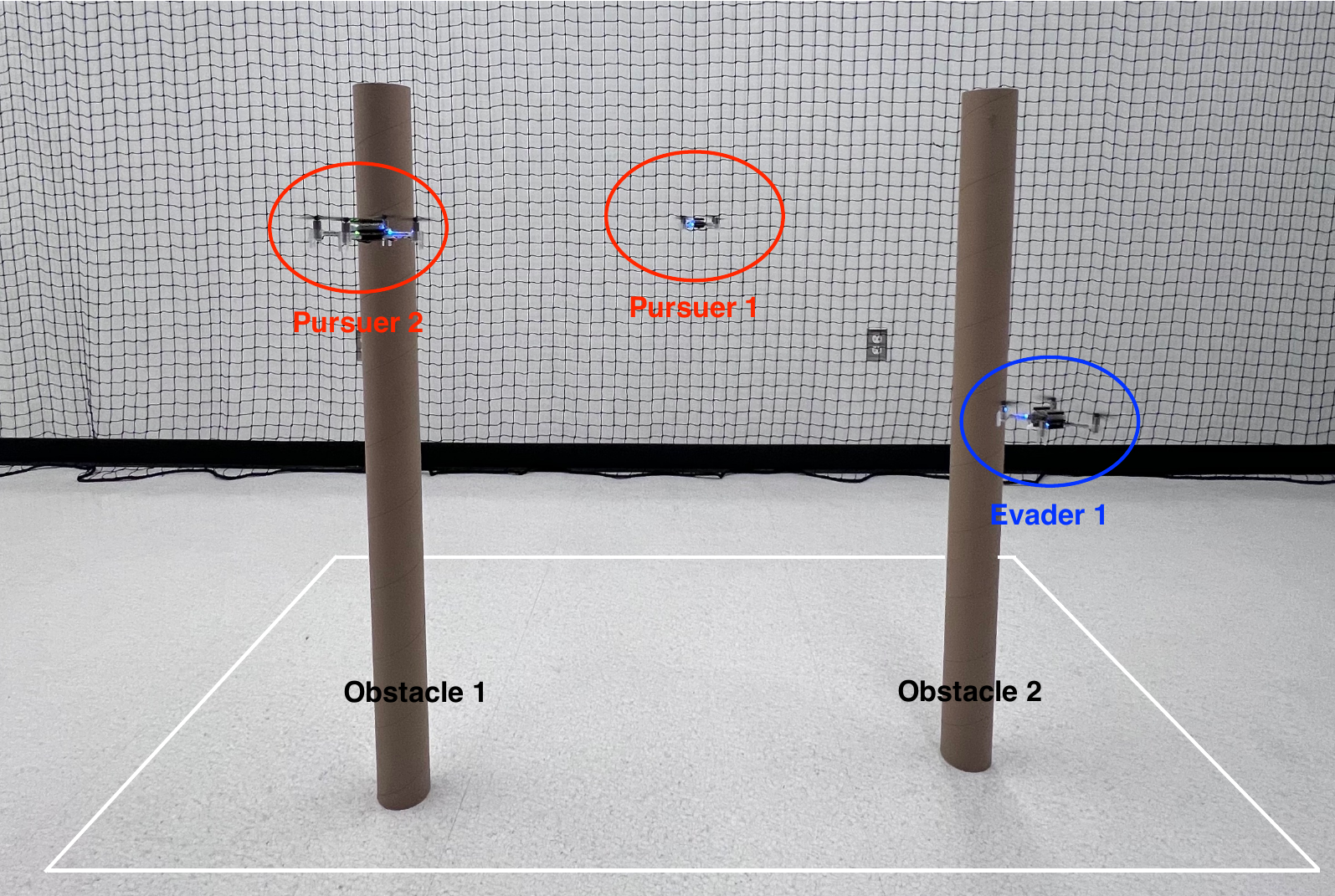}
    \caption{The real-world experimental setup for the \textit{Tag-occlusion} task, featuring two Crazyflie drones as pursuers (red) and one as an evader (blue), navigating around two physical obstacles.}
    \label{fig:experiment_setup}
    % \vspace{-10pt}
\end{figure}

\textbf{Experimental setup.}
We deploy a LEGO policy trained in simulation on a physical
\emph{Tag-occlusion} setup with two pursuers, one evader, and two
obstacles, as shown in Figure~\ref{fig:experiment_setup}. Each agent is
implemented using a Crazyflie 2.1+ nano quadrotor. The initial drone
poses are measured manually to establish a common global frame, after
which onboard IMU measurements and Kalman-filtered state estimates are
used to track the drones and construct the global state $X$. The policy
is deployed without additional real-world training.

For safe operation, the pursuers and evader fly at different altitudes
to avoid collisions at capture. A drone is briefly paused when another
approaches too closely to reduce downwash effects. We also use obstacles
with diameter $0.1\,\mathrm{m}$, slightly smaller than those used during
training, to provide an additional safety margin.

% \textbf{Experimental Setup.} In the real world, we demonstrate our LEGO approach in the \emph{Tag-occlusion} environment consisting of two pursuers, one evader, and two obstacles, as illustrated in Figure~\ref{fig:experiment_setup}. Both the pursuers and the evader are realized with \emph{Crazyflie 2.1+} nano drones. 
% % The position of each drone is tracked using its built-in IMU with Kalman filtering~\cite{welch1995introduction}, which is used to construct the global state $X$.
% \change{The initial pose of each drone is measured manually to establish a common global frame, after which the on-board IMU tracks each drone's motion via Kalman filtering~\cite{welch1995introduction} to form the global state $X$.}
% We first train our policy in simulation and then deploy it on the real-world platform.

\begin{figure}
    \centering
    % \vspace{-10pt}
    \includegraphics[width=\linewidth]{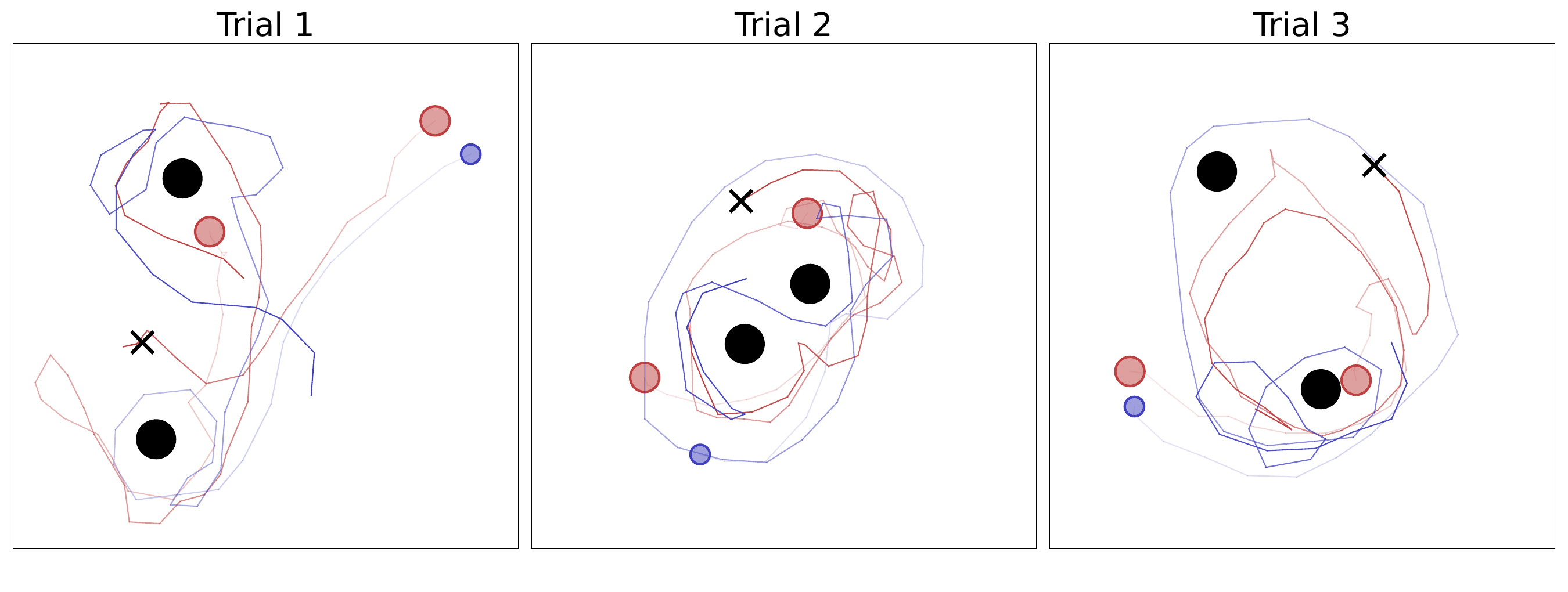}
    \caption{Trajectories from three real-world trials with one pursuer
disabled during execution. $\times$ marks the pursuer's position at
failure; the remaining pursuer continues to chase the evader.}
    \label{fig:trajectory_demo}
    % \vspace{-10pt}
\end{figure}

% In simulation, once the pursuers capture the evader, both agents can continue moving without consequence. In the real world, however, such a capture would result in an unavoidable collision, potentially damaging the drones and causing the experiment to fail. To prevent this, the pursuers and evader are assigned to maneuver at different altitudes. To prevent downwash effects, one drone is briefly paused when it approaches another too closely. Furthermore, to reduce the risk of collisions with obstacles, we use obstacles with a diameter of 0.1\,m, which is slightly smaller than the size used during policy training.
\textbf{Robustness to agent failure.}
We evaluate deployment under an unexpected reduction in team size using
trials of horizon $T=100$. At time $t=30$, one pursuer is disabled
and remains stationary, without retraining or fine-tuning the policy.
In Figure~\ref{fig:trajectory_demo}, $\times$ marks the pursuer's
position at failure. The remaining pursuer continues to pursue the
evader, demonstrating that LEGO remains operational after the loss of
one agent. Qualitatively, the active pursuer transitions from coordinated
blocking to direct pursuit, while the stationary pursuer continues to
constrain the evader's motion.
% \textbf{Robustness.} To demonstrate that zero-shot scalability improves robustness in real-world settings, we design a scenario with a total horizon of $T=100$ in which one of the pursuers lands off at the time-step $t=30$, (i.e., breaks down during the chasing). As shown in Figure~\ref{fig:trajectory_demo}, where $\times$ marks the position at which one pursuer breaks down, while the remaining pursuer continues to chase the evader, highlighting that our method maintains functionality even under agent failure. Specifically, although the "broken" pursuer is inactive, it serves as a roadblock that prevents the evader from approaching its vicinity, while the remaining pursuer switches its strategy from blocking the evader’s path to directly chasing it.

% We design a set of experiments to validate three key claims: 1) Incorporating geometric and permutation equivariance improves training sample efficiency, thereby enhancing overall learning performance; 2) The proposed LEGO framework is versatile, applicable to a wide range of multi-agent tasks, from cooperative to competitive settings; 3) Equivariance with Graph Neural Network structure indeed improve generalization performance.

% \begin{figure}
%     \centering
%     \includegraphics[width=\linewidth]{Figures/generalization_rewards.pdf}
%     \caption{\textbf{Average evaluation reward result (the higher the better).}  For LEGO-MAPPO, a single policy is trained only in the MPE Spread environment with 4 agents and 4 landmarks. In contrast, MAPPO is trained separately under each setting. 
%     % $\dagger$ denotes models that are specifically trained for the corresponding setting.
%     }
%     \label{fig:generalization_rewards}
% \end{figure}

\section{Conclusion}

In this work, we introduced Local-Canonicalization Equivariant Graph
Neural Networks (LEGO), a modular MARL architecture that combines
agent-centric canonicalization with role-aware graph encoding to produce
$E(2)$- and intra-role permutation-equivariant policies. Across
cooperative and competitive benchmarks, LEGO-MAPPO improves sample
efficiency and task performance relative to standard, graph-based, and
equivariant baselines. The learned policies transfer without fine-tuning
to unseen team sizes, maintain performance under spatial distribution
shifts, and remain operational on Crazyflie hardware after one pursuer
is disabled. Future work will extend LEGO to three-dimensional swarm
control, where constructing stable equivariant canonical frames is a
central challenge~\cite{lippmann2024beyond}.

\renewcommand*{\bibfont}{\footnotesize}
\bibliographystyle{IEEEtran}
\bibliography{ref}

\end{document}